\documentclass{article}

\usepackage{PRIMEarxiv}

\usepackage[utf8]{inputenc} 
\usepackage[T1]{fontenc}    
\usepackage{hyperref}       
\usepackage{url}            
\usepackage{booktabs}       
\usepackage{amsfonts}       
\usepackage{nicefrac}       
\usepackage{microtype}      
\usepackage{lipsum}
\usepackage{fancyhdr}       
\usepackage{graphicx}       
\graphicspath{{media/}}     

\title{VIRAL: Vision-grounded Integration \\ for Reward design And Learning
}

\author{
  Valentin Cuzin-Rambaud, Emilien Komlenovic, Alexandre Faure, Bruno Yun\\
  Université Claude Bernard Lyon 1, France \\
  CNRS, Ecole Centrale de Lyon, \\
  INSA Lyon, Université Lumière Lyon 2,\\
  LIRIS, UMR5205, France\\
  \texttt{\{valentin.cuzin-rambaud,bruno.yun\}@univ-lyon1.fr} \\
  \texttt{emilien.komlenovic.prow@gmail.com} \\
}

\begin{document}
\maketitle

\begin{abstract}
  The alignment between humans and machines is a critical challenge in artificial intelligence today. Reinforcement learning, which aims to maximize a reward function, is particularly vulnerable to the risks associated with poorly designed reward functions. Recent advancements has shown that Large Language Models (LLMs) for reward generation can outperform human performance in this context. We introduce VIRAL, a pipeline for generating and refining reward functions through the use of multi-modal LLMs. VIRAL autonomously creates and interactively improves reward functions based on a given environment and a goal prompt or annotated image.

The refinement process can incorporate human feedback or be guided by a description generated by a video LLM, which explains the agent's policy in video form. 
We evaluated VIRAL in five Gymnasium environments, demonstrating that it accelerates the learning of new behaviors while ensuring improved alignment with user intent. 
The source-code and demo video are available at: \url{https://github.com/VIRAL-UCBL1/VIRAL} and \url{https://youtu.be/BBIB4bEVSjE}.
\end{abstract}


\maketitle

\section{Introduction}

Reward shaping \cite{DBLP:conf/icml/NgHR99} is a fundamental challenge in Reinforcement Learning (RL), involving the design of reward functions that efficiently guide an agent towards desired behaviors. 
%
%
A poorly designed reward can lead to unintended behaviors, while a well-crafted one accelerates learning and ensures alignment with human intent. However, designing effective rewards is labor-intensive and requires significant expertise, particularly in complex environments \cite{DBLP:conf/nips/ChristianoLBMLA17,DBLP:conf/nips/IbarzLPILA18,DBLP:conf/icml/LeeSA21,DBLP:conf/iclr/ParkSSLAL22}. 

Early work on automated reward design \cite{nlp_reward_2019} leveraged natural language processing techniques — such as recurrent neural networks and word embeddings — to construct reward signals, demonstrating promising results in ATARI games. 
More recently, LLMs have gained traction in RL and robotics due to their versatility in solving diverse problems \cite{DBLP:conf/corl/IchterBCFHHHIIJ22,DBLP:conf/icra/SinghBMGXTFTG23}. 
Early attempts at using LLMs for reward shaping \cite{kwon2023reward} employed GPT-3 as a binary reward signal, but this approach was limited in scope and applicability. More recent advances \cite{ma2023eureka, song2023self, xietext2reward} have leveraged OpenAI's GPT-4 model to generate code for reward functions, demonstrating competitive performance and improved adaptability across different environments.
However, these methods only focus on the text-based inputs (disregarding vision) and their reliance on a closed-source, and computationally expensive LLM for performance, hinders reproducibility and accessibility.

We propose a \textbf{\underline{V}}ision-grounded \textbf{\underline{I}}ntegration for \textbf{\underline{R}}eward design \textbf{\underline{A}}nd \textbf{\underline{L}}earning (VIRAL) framework to design rewards functions from simple users prompts and/or annotated image.
VIRAL sets itself from the state-of-the-art \cite{ma2023eureka, xietext2reward} in several key ways. 
First, VIRAL prioritizes the use of open-source, efficient, and lightweight LLMs \cite{hui2024qwen2,guo2025deepseek}, ensuring greater accessibility, cost efficiency, and transparency.
Second, unlike prior methods, VIRAL integrates Large Vision Language Models (LVLMs) \cite{liu2024visual, meta-llama/Llama-3.2-11B-Vision-Instruct} to process both text and images, enhancing its ability to interpret user intent more accurately. 
Third, VIRAL is the first reward shaping approach to incorporate Video-LVLMs \cite{lin2023video} for describing the movements of objects within the scene, providing richer context for reward function generation.
Finally, instead of relying on direct access to the environment's code (as in EUREKA \cite{ma2023eureka}) or structured abstraction (through Pydantic class definitions as in Text2Reward \cite{xietext2reward}), VIRAL describes environments solely through its observations, following the Gymnasium \cite{towers2024gymnasium} documentation. This simplifies implementation for users while ensuring the LLM captures the necessary information for coherent reward generation. 
Our main contributions are as follows: 

\begin{itemize}
    \item The VIRAL pipeline to automatically design reward functions using a simple natural language prompt and/or annotated image.
    \item A self-refinement process for reward functions, augmented with human feedback or video description from a LVLM.
    \item A scalable and modular implementation designed to adapt to various RL problems within the Gymnasium framework, leveraging multiprocessing for faster inference.
    \item An evaluation of VIRAL in five Gymnasium environments, showing its various benefits for learning and user intent alignment, using an empirical study.
\end{itemize}

This paper is organized as follows: Section \ref{section:architecture} details the architecture and inner workings of VIRAL. Section \ref{sec:eval} presents our evaluation methodology and results. Section \ref{sec:conclusion} provides concluding remarks. 

\section{The VIRAL Architecture}
\label{section:architecture}

%

%

This section details the VIRAL procedure to generate multiple rewards functions which are rated to find the best agent behaviors.

\begin{figure*}[h!]
    \centering
    \includegraphics[width=0.9\textwidth]{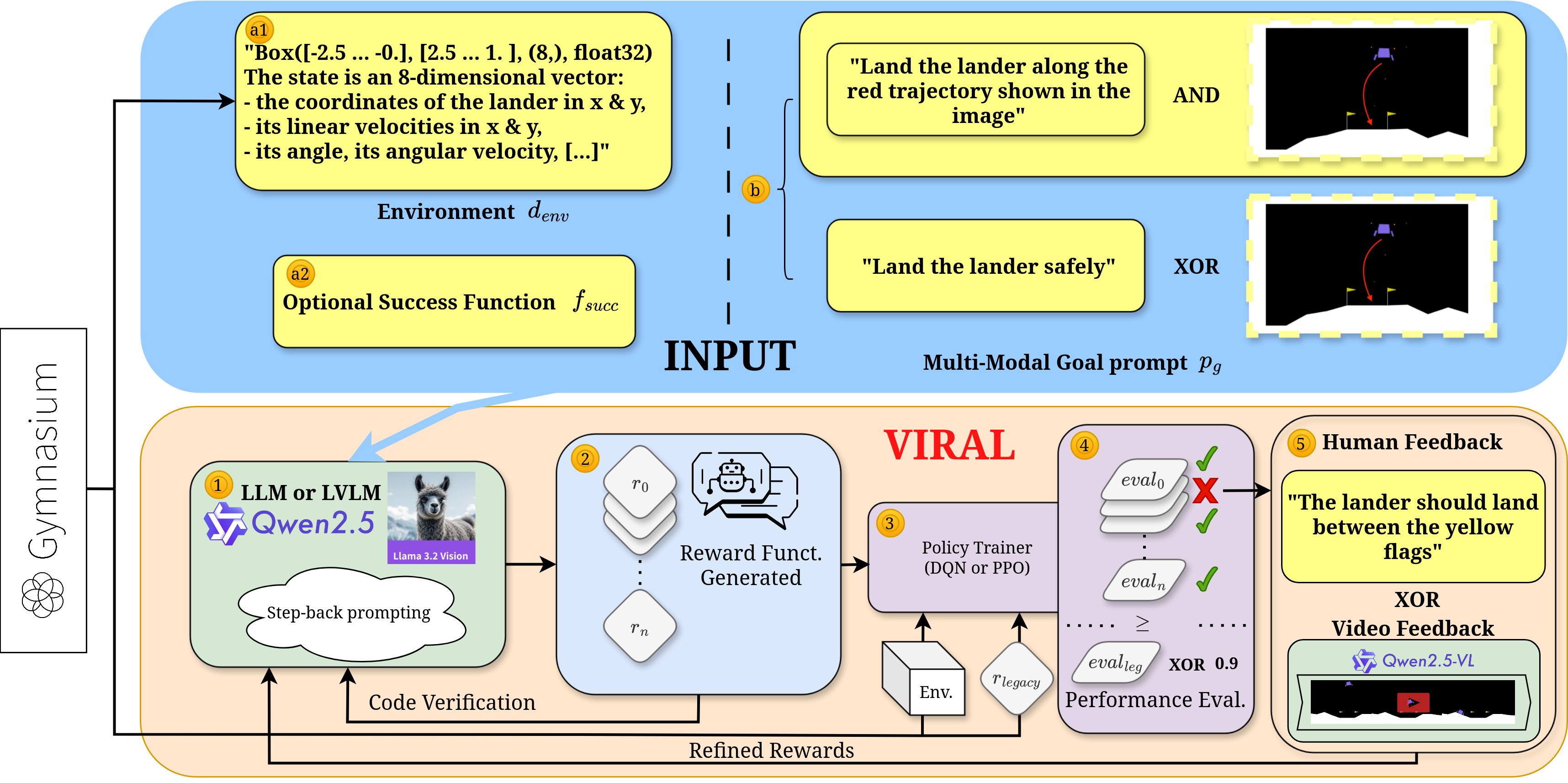}
    \caption{The VIRAL pipeline. Given an input (a textual environment description, an optional success function, and a goal prompt), the system generates a set of reward functions and iteratively refines them.}
    \label{fig:pipeline}
\end{figure*}

\subsection{The Input Parameters}

For its input, VIRAL uses a set of specific elements (see top of Figure~\ref{fig:pipeline}). 
It includes ($a_1$) a textual environment description $d_{env}$, ($a_2$) an optional implementation of a success function $f_{succ}$ provided as Python code, and $(b)$ the goal prompt $p_g$ (which can be multi-modal).


To give the prompt $p_g$, the user has the possibility to choose between a text prompt, an annotated image $img$, or both of those. 
The $img$ can include annotations (with arrows, text, areas, etc.). 
The input $d_{env}$ is extracted from Gymnasium and provides an accurate representation of the environment's observable space to the LLM, adding more depth to the generated reward function. Using text for $d_{env}$ allows for any Gymnasium environment to be seamlessly integrated with VIRAL. The input $f_{succ}: States(env) \to \{0,1\}$ is a success function which must be tailored to the goal. This function determines how success is manifested within the specific context of the task, helping the LLM in designing the reward function. Note that while $f_{succ}$ must be related to the goal prompt $p_g$, returning $1$ (success) can mean a partial success. For example, for a textual goal prompt $p_g:$ ``Land the lander along the red trajectory shown in the image'', $f_{succ}$ may return $1$ if the lander managed to land, ignoring the trajectory.


\subsection{The Initial Generation}

For the initial generation, we opted for zero-shot prompting, despite the effectiveness of few-shot prompting \cite{brown2020language}. This was motivated by generalization needs and the difficulty for users to provide examples, making zero-shot prompting more practical and user-friendly.
%


VIRAL implements a collaboration between two LLMs (critic and coder), with specific system prompts for their roles. The former must be a good supervisor and specify steps to help the latter in producing good quality code. 
Among the strategies employed to enhance zero-shot generation, one particularly effective method is step-back prompting \cite{zheng2023take}, which allows to obtain a broader perspective by first reasoning about a problem at a higher-level before generating a detailed response (see step 1 of Fig. \ref{fig:pipeline}).
The critic LLM generates the step-back prompt which is given to the coder LLM (see step 2 of Fig. \ref{fig:pipeline}). Note that only the critic LLM needs to be multi-modal.
%

The generated code is checked for syntax and logical issues by using a try-catch process. If errors are caught, they are sent back to the coder LLM via a custom prompt, allowing it to revise its output. 

\subsection{Learning a Policy}

The generated reward functions take an observation as parameters, along with two boolean values indicating whether there is a success or failure (note that an agent can be in a neutral state, without a success or a failure). Once a reward function is generated, an RL algorithm is used to make the agent search for its policy. We restrict ourselves to Deep Q-Network (DQN) \cite{mnihPlayingAtariDeep2013} and Proximal Policy Optimization (PPO) \cite{schulmanProximalPolicyOptimization2017} 
 and select the most suitable one for each environment (step 3 of Fig. \ref{fig:pipeline}).
%
During training, rewards and observations are retrieved, and if the user wishes to, they can implement objective metric functions that take these observations and return a dictionary of useful objective metrics for this environment. During testing we can compare the success rate of our custom reward, with the baseline ``legacy" reward function $r_{legacy}$ or with a user-defined threshold, which determines an acceptable success rate (step 4 of Fig. \ref{fig:pipeline}).



\subsection{The Refinement Process}

The refining process unfolds as follows (see step 5 of Fig. \ref{fig:pipeline}).



The critic LLM analyzes the training results by leveraging collected statistics like state-observations during the run, and an optional user or Video-LVLM feedback. 
Indeed, to identify potential reasons why the reward function underperformed, a user or Qwen2.5-VL feedback can provide a more precise description of the agent's learned behavior. 
These observations are passed to the coder LLM, which creates an improved reward function by addressing the identified weaknesses and aligning with the intended objectives

These methods are combined into an iterative approach. In each iteration, a refined reward function is evaluated and compared to its previous version.
If it does not pass the evaluation, the refinement process is repeated until the desired performance level is achieved.


\section{Empirical evaluation}
\label{sec:eval}

For our evaluation, we used \textit{Qwen2.5-Coder-32B} \cite{qwen2,hui2024qwen2} as the coder LLM due to its performance being comparable to that of \textit{GPT-4}, making it a reliable choice for this role. For the multi-modal critic LLM, we chose \textit{Llama3.2-Vision-11B} \cite{meta-llama/Llama-3.2-11B-Vision-Instruct}, as it is lightweight, open-source, and well-suited for our framework’s requirements. For Video-LVLM, we used \textit{Qwen2.5-VL-7B} \cite{Qwen2VL,qwen2.5-VL}. However, we note that our framework is model-agnostic and allows the use of any LLM for the coder and critic LLMs and Video-LVLM. The evaluation was performed with a NVIDIA A40 GPU and a 12-cores Intel CPU.

In our experiments, we used a selection of environments from the Gymnasium toolkit to evaluate the performance of our generated reward functions. Namely, the classic environments of \textit{CartPole} and \textit{Lunar Lander} allowed us to test fundamental control and optimization strategies in simple yet challenging scenarios. Next, we selected the \textit{Highway} environment, where a vehicle must navigate a multi-lane road, avoiding collisions and optimizing its speed while adhering to driving rules. Given the increasing relevance of autonomous vehicles, we considered it essential to include this environment in our study as it addresses modern-day challenges in automation and decision-making. Finally, we incorporated two robotics-focused environments, \textit{Hopper} and \textit{Swimmer}, both derived from the MuJoCo physics simulator. These environments played a crucial role in evaluating the efficiency of the generated reward function for robotic locomotion and control systems.

All results are available in CSV format in our GitHub repository.


\paragraph{Better Rewards.}

We evaluated the efficiency of VIRAL by comparing the behavior of agents trained using our generated reward function (without the refining process) against those trained with the legacy reward function. 
We instantiated VIRAL with Qwen2.5-coder-32B as the critic/coder LLM (text-only goal prompt).

For the CartPole environment trained with PPO, and the goal prompt 
\textit{"Create a reward function for the CartPole environment that encourages keeping the pole upright for as long as possible."}, Table~\ref{tab:cartpole_compare} shows that the policy learned with our reward function outperforms that of Gymnasium. We limited the number of epochs to 25 000 in order to make the learning problem more complex and suitable for evaluating learning speed. Additionally, we can see that the cart moves slightly more on average, resulting in a more stable pole overall.

%


    
        
        

\begin{table}[ht]
\caption{Comparison of legacy reward vs. ours  on Cartpole Environment(avg. over 10 runs)}
\label{tab:cartpole_compare}
\centering
\begin{tabular}{llll}
Reward function & Cart position diff. & Pole angle diff. & Success rate \\
\toprule
gymnasium & 0.2812$\pm$0.0012 & 0.0645$\pm$0.0232 & 0.5870$\pm$0.2384 \\
ours & 0.3081$\pm$0.0277 & 0.0624$\pm$0.0019 & \textbf{0.8530$\pm$0.1864} \\
\bottomrule
\end{tabular}
\end{table}

For the \textit{Highway} environment trained with DQN, and the goal prompt \textit{"Control the ego vehicle to reach a high speed without collision."}, we have better success rates than the legacy reward function 7 times out of 10 (with a median success rate of 0.78).


\paragraph{Semantic Alignment.}

We conducted a study with 25 annotators to determine whether the learned behavior is aligned with the goal prompt provided (text, image or both). For a full description of the goal prompt used, we refer the reader to our Github repository. Each annotator was tasked to annotate a maximum of 120 videos (4 environments and 10 videos per goal prompt modality). On average, each annotator rated 71 videos for a total of 1777 annotated videos. 
For each video, annotators used 5-point Likert scale (from ``Strongly disagree'' to ``Strongly agree'') to answer the following two questions:
(1) \textit{I understand the instructions}, and 
(2) \textit{The instructions are followed}.
The human evaluation allowed us to assess the effectiveness of the modality on the semantic alignment.

For the first item, we obtained a mean of $4.89\pm0.41$ and with only 147 (out of 1777) videos with an understanding score less than 5/5, highlighting that very few goal prompts were misunderstood, confirming that goal prompts were well-crafted and understandable. This indicates that the study's results are therefore not confounded by ambiguity in the prompts themselves.

\begin{figure}
    \centering
    \includegraphics[width=0.9\linewidth]{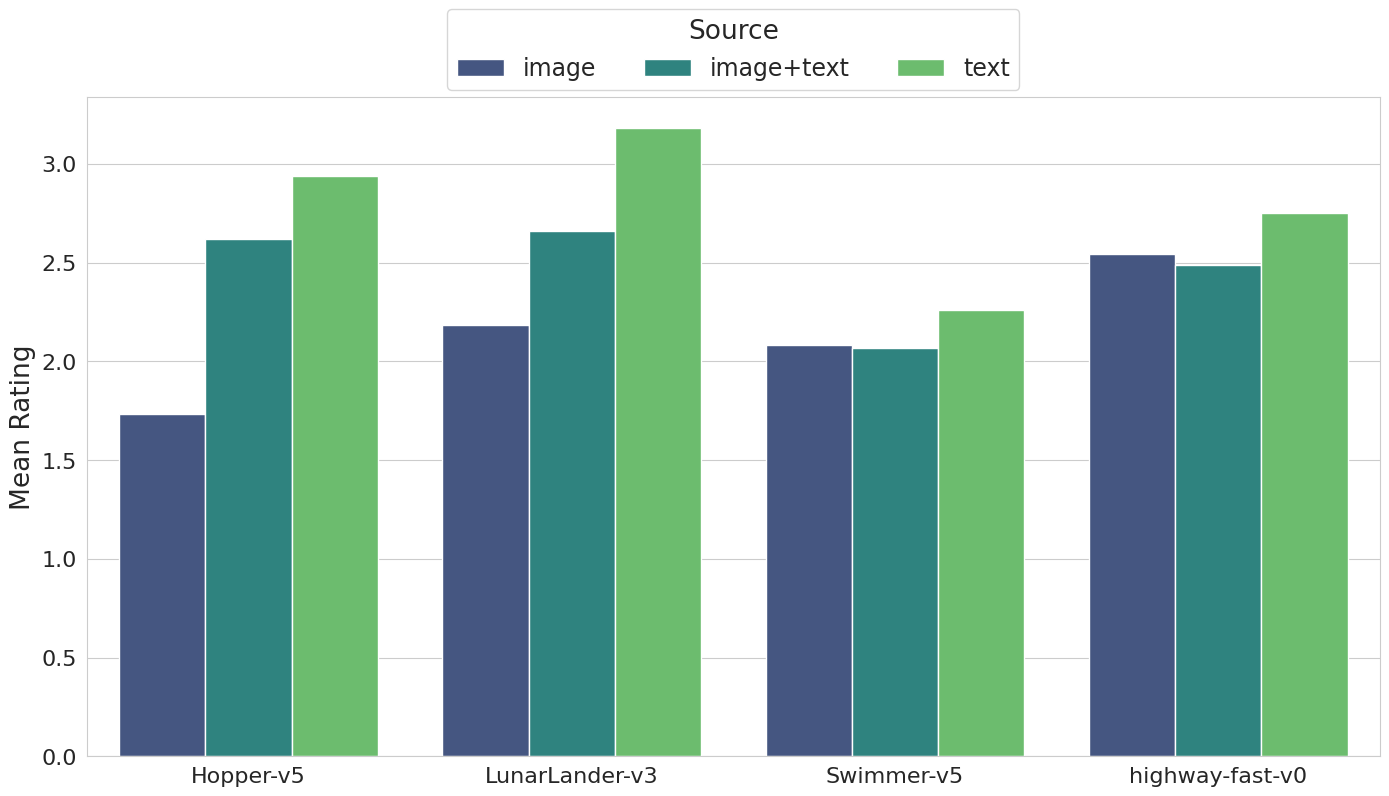}
    \caption{Semantic alignment over the 10 videos of each annotator for different gymnasium's environments and modalities.}
    \label{fig:mean_rating}
\end{figure}

Fig. \ref{fig:mean_rating} shows the mean ratings for the second item  (over the 10 videos of each annotator) for each environment and modality.
We note that text-only prompts resulted in the highest semantic alignment, indicating that textual descriptions provided clearer, more actionable guidance for the system than other modalities.
Overall, although using an image-only prompt was not as good as a text-only goal prompt, we still obtained a mean of $2.14\pm1.42$ (out of 5) across environments for the image modality.
Adding text to image goal prompts did not consistently improve alignment, and in some cases hurt performance (increase in Hopper and LunarLander environments and decrease in the Swimmer and Highway-fast environments), suggesting potential interference or confusion introduced by multimodal inputs.
The relative performance of each modality varied with the environment, which implies that the usefulness of multimodal or visual instructions may depend heavily on the task's nature.

However, we emphasize that VIRAL's primary objective is to discover the optimal reward across multiple generations. Therefore, we argue that maximizing semantic alignment in the best-performing instances is more relevant than focusing solely on the average alignment across all videos for a given modality.
Specifically, for each video, we compute the average semantic alignment score reported by the annotators, and then, for each modality, we report the highest of these averages across the 10 videos in Table~\ref{tab:avg_rating}.
We can make the following three observations: \textbf{(1) Textual goal prompts generally lead to the best alignment.} Indeed, in the Hopper environment, the textual modality achieves the highest score $(4.92 \pm 0.28)$, outperforming both image and image+text. Moreover, in the other environments, the textual modality is nearly tied to the best modality. This suggests that textual goal prompts alone are highly effective at conveying the intended behavior across diverse environments.
\textbf{(2) Image-only goal prompts can occasionally perform best (in the Swimmer and Highway-fast environments), but are less consistent.} 
This indicates that visual cues alone can be sufficient, but their effectiveness is highly environment-dependent — likely tied to how interpretable or informative the visual goal is for that task.
Moreover, the variance is relatively high which could imply greater inconsistency in human judgments when interpreting the alignment with image-only goal prompts.
\textbf{(3) Text+image goal prompts can improve alignment but does not improve it consistently.} 
While image+text performs well in LunarLander ($4.93\pm0.26$), it underperforms in the other modalities.
This reinforces our previous finding that adding text to an image goal prompt does not always help, and might even introduce noise or ambiguity in some environments.



\begin{table}[]
    \centering
    \caption{Maximum Average Semantic Alignment per Modality and Environment.}
    \begin{tabular}{lccc}
        Environment & Image $p_g$ & Textual $p_g$ & Image+Textual $p_g$\\
        \midrule
        Hopper-v5 & 3.41$\pm$1.00 & \textbf{4.92$\pm$0.28} & 4.83$\pm$0.39 \\
        LunarLander-v3 & 3.83$\pm$1.26 & 4.92$\pm$0.28 & \textbf{4.93$\pm$0.26} \\
        Swimmer-v5 & \textbf{4.19$\pm$0.75} & 4.14$\pm$0.95 & 3.41$\pm$1.33 \\
        Highway-fast-v0 & \textbf{4.44$\pm$1.29} & 4.06$\pm$1.03 & 3.91$\pm$1.38 \\
    \end{tabular}
    \label{tab:avg_rating}
\end{table}



\paragraph{Enhanced with Feedback.}

To show the improvement of a generated reward function brought by the refining process, we can compare its performance to the one of a reward function after a single iteration of feedback obtained by Qwen2.5-VL-7B. 

In the LunarLander environment and defining the success criterion as a safe landing between the two yellow poles, 
we carried out 10 runs, with and without feedback and 100 tests at the end of each training session. We found that the use of the Video-LVLM creates reward functions that are on average 18.33\% better based on the success rate.

\section{Conclusion}
\label{sec:conclusion}

We introduced VIRAL, a pipeline for generating and refining reward functions through the use of multi-modal LLMs.
We demonstrated that our approach outperformed legacy reward functions in terms of performance and flexibility. Our results showed that agents were able to learn new behaviors based on simplistic sketches, highlighting the robustness of our approach. Furthermore, the integration of feedback (from a Video-LVLM or a user) enabled agents to better align with the intended objectives, providing a deeper understanding of the desired behaviors.

In future work, we plan to adapt pre-existing policies to learn new behaviors. This approach could pave the way for greater policy generalization and smoother transitions between different sets of complex tasks.

\bibliographystyle{unsrt}  
\bibliography{mybibfile}

\begin{thebibliography}{10}

\bibitem{DBLP:conf/icml/NgHR99}
Andrew~Y. Ng, Daishi Harada, and Stuart Russell.
\newblock Policy invariance under reward transformations: Theory and
  application to reward shaping.
\newblock In Ivan Bratko and Saso Dzeroski, editors, {\em Proceedings of the
  Sixteenth International Conference on Machine Learning {(ICML} 1999), Bled,
  Slovenia, June 27 - 30, 1999}, pages 278--287. Morgan Kaufmann, 1999.

\bibitem{DBLP:conf/nips/ChristianoLBMLA17}
Paul~F. Christiano, Jan Leike, Tom~B. Brown, Miljan Martic, Shane Legg, and
  Dario Amodei.
\newblock Deep reinforcement learning from human preferences.
\newblock In Isabelle Guyon, Ulrike von Luxburg, Samy Bengio, Hanna~M. Wallach,
  Rob Fergus, S.~V.~N. Vishwanathan, and Roman Garnett, editors, {\em Advances
  in Neural Information Processing Systems 30: Annual Conference on Neural
  Information Processing Systems 2017, December 4-9, 2017, Long Beach, CA,
  {USA}}, pages 4299--4307, 2017.

\bibitem{DBLP:conf/nips/IbarzLPILA18}
Borja Ibarz, Jan Leike, Tobias Pohlen, Geoffrey Irving, Shane Legg, and Dario
  Amodei.
\newblock Reward learning from human preferences and demonstrations in atari.
\newblock In Samy Bengio, Hanna~M. Wallach, Hugo Larochelle, Kristen Grauman,
  Nicol{\`{o}} Cesa{-}Bianchi, and Roman Garnett, editors, {\em Advances in
  Neural Information Processing Systems 31: Annual Conference on Neural
  Information Processing Systems 2018, NeurIPS 2018, December 3-8, 2018,
  Montr{\'{e}}al, Canada}, pages 8022--8034, 2018.

\bibitem{DBLP:conf/icml/LeeSA21}
Kimin Lee, Laura~M. Smith, and Pieter Abbeel.
\newblock {PEBBLE:} feedback-efficient interactive reinforcement learning via
  relabeling experience and unsupervised pre-training.
\newblock In Marina Meila and Tong Zhang, editors, {\em Proceedings of the 38th
  International Conference on Machine Learning, {ICML} 2021, 18-24 July 2021,
  Virtual Event}, volume 139 of {\em Proceedings of Machine Learning Research},
  pages 6152--6163. {PMLR}, 2021.

\bibitem{DBLP:conf/iclr/ParkSSLAL22}
Jongjin Park, Younggyo Seo, Jinwoo Shin, Honglak Lee, Pieter Abbeel, and Kimin
  Lee.
\newblock {SURF:} semi-supervised reward learning with data augmentation for
  feedback-efficient preference-based reinforcement learning.
\newblock In {\em The Tenth International Conference on Learning
  Representations, {ICLR} 2022, Virtual Event, April 25-29, 2022}.
  OpenReview.net, 2022.

\bibitem{nlp_reward_2019}
Prasoon Goyal, Scott Niekum, and Raymond~J Mooney.
\newblock Using natural language for reward shaping in reinforcement learning.
\newblock {\em arXiv preprint arXiv:1903.02020}, 2019.

\bibitem{DBLP:conf/corl/IchterBCFHHHIIJ22}
Brian Ichter, Anthony Brohan, Yevgen Chebotar, Chelsea Finn, Karol Hausman,
  Alexander Herzog, Daniel Ho, Julian Ibarz, Alex Irpan, Eric Jang, Ryan
  Julian, Dmitry Kalashnikov, Sergey Levine, Yao Lu, Carolina Parada, Kanishka
  Rao, Pierre Sermanet, Alexander Toshev, Vincent Vanhoucke, Fei Xia, Ted Xiao,
  Peng Xu, Mengyuan Yan, Noah Brown, Michael Ahn, Omar Cortes, Nicolas Sievers,
  Clayton Tan, Sichun Xu, Diego Reyes, Jarek Rettinghouse, Jornell Quiambao,
  Peter Pastor, Linda Luu, Kuang{-}Huei Lee, Yuheng Kuang, Sally Jesmonth,
  Nikhil~J. Joshi, Kyle Jeffrey, Rosario~Jauregui Ruano, Jasmine Hsu, Keerthana
  Gopalakrishnan, Byron David, Andy Zeng, and Chuyuan~Kelly Fu.
\newblock Do as {I} can, not as {I} say: Grounding language in robotic
  affordances.
\newblock In Karen Liu, Dana Kulic, and Jeffrey Ichnowski, editors, {\em
  Conference on Robot Learning, CoRL 2022, 14-18 December 2022, Auckland, New
  Zealand}, volume 205 of {\em Proceedings of Machine Learning Research}, pages
  287--318. {PMLR}, 2022.

\bibitem{DBLP:conf/icra/SinghBMGXTFTG23}
Ishika Singh, Valts Blukis, Arsalan Mousavian, Ankit Goyal, Danfei Xu, Jonathan
  Tremblay, Dieter Fox, Jesse Thomason, and Animesh Garg.
\newblock Progprompt: Generating situated robot task plans using large language
  models.
\newblock In {\em {IEEE} International Conference on Robotics and Automation,
  {ICRA} 2023, London, UK, May 29 - June 2, 2023}, pages 11523--11530. {IEEE},
  2023.

\bibitem{kwon2023reward}
Minae Kwon, Sang~Michael Xie, Kalesha Bullard, and Dorsa Sadigh.
\newblock Reward design with language models.
\newblock {\em arXiv preprint arXiv:2303.00001}, 2023.

\bibitem{ma2023eureka}
Yecheng~Jason Ma, William Liang, Guanzhi Wang, De-An Huang, Osbert Bastani,
  Dinesh Jayaraman, Yuke Zhu, Linxi Fan, and Anima Anandkumar.
\newblock Eureka: Human-level reward design via coding large language models.
\newblock {\em arXiv preprint arXiv:2310.12931}, 2023.

\bibitem{song2023self}
Jiayang Song, Zhehua Zhou, Jiawei Liu, Chunrong Fang, Zhan Shu, and Lei Ma.
\newblock Self-refined large language model as automated reward function
  designer for deep reinforcement learning in robotics.
\newblock {\em arXiv preprint arXiv:2309.06687}, 2023.

\bibitem{xietext2reward}
Tianbao Xie, Siheng Zhao, Chen~Henry Wu, Yitao Liu, Qian Luo, Victor Zhong,
  Yanchao Yang, and Tao Yu.
\newblock Text2reward: Reward shaping with language models for reinforcement
  learning.
\newblock In {\em The Twelfth International Conference on Learning
  Representations}, 2024.

\bibitem{hui2024qwen2}
Binyuan Hui, Jian Yang, Zeyu Cui, Jiaxi Yang, Dayiheng Liu, Lei Zhang, Tianyu
  Liu, Jiajun Zhang, Bowen Yu, Kai Dang, et~al.
\newblock Qwen2.5-coder technical report.
\newblock {\em arXiv preprint arXiv:2409.12186}, 2024.

\bibitem{guo2025deepseek}
Daya Guo, Dejian Yang, Haowei Zhang, Junxiao Song, Ruoyu Zhang, Runxin Xu,
  Qihao Zhu, Shirong Ma, Peiyi Wang, Xiao Bi, et~al.
\newblock Deepseek-r1: Incentivizing reasoning capability in llms via
  reinforcement learning.
\newblock {\em arXiv preprint arXiv:2501.12948}, 2025.

\bibitem{liu2024visual}
Haotian Liu, Chunyuan Li, Qingyang Wu, and Yong~Jae Lee.
\newblock Visual instruction tuning.
\newblock {\em Advances in neural information processing systems}, 36, 2024.

\bibitem{meta-llama/Llama-3.2-11B-Vision-Instruct}
Meta.
\newblock Llama-3.2-11b-vision-instruct.
\newblock
  \url{https://huggingface.co/meta-llama/Llama-3.2-11B-Vision-Instruct}, 2024.

\bibitem{lin2023video}
Bin Lin, Yang Ye, Bin Zhu, Jiaxi Cui, Munan Ning, Peng Jin, and Li~Yuan.
\newblock Video-llava: Learning united visual representation by alignment
  before projection.
\newblock {\em arXiv preprint arXiv:2311.10122}, 2023.

\bibitem{towers2024gymnasium}
Mark Towers, Ariel Kwiatkowski, Jordan Terry, John~U Balis, Gianluca De~Cola,
  Tristan Deleu, Manuel Goul{\~a}o, Andreas Kallinteris, Markus Krimmel, Arjun
  KG, et~al.
\newblock Gymnasium: A standard interface for reinforcement learning
  environments.
\newblock {\em arXiv preprint arXiv:2407.17032}, 2024.

\bibitem{brown2020language}
Tom Brown, Benjamin Mann, Nick Ryder, Melanie Subbiah, Jared~D Kaplan, Prafulla
  Dhariwal, Arvind Neelakantan, Pranav Shyam, Girish Sastry, Amanda Askell,
  et~al.
\newblock Language models are few-shot learners.
\newblock {\em Advances in neural information processing systems},
  33:1877--1901, 2020.

\bibitem{zheng2023take}
Huaixiu~Steven Zheng, Swaroop Mishra, Xinyun Chen, Heng-Tze Cheng, Ed~H Chi,
  Quoc~V Le, and Denny Zhou.
\newblock Take a step back: Evoking reasoning via abstraction in large language
  models.
\newblock {\em arXiv preprint arXiv:2310.06117}, 2023.

\bibitem{mnihPlayingAtariDeep2013}
Volodymyr Mnih, Koray Kavukcuoglu, David Silver, Alex Graves, Ioannis
  Antonoglou, Daan Wierstra, and Martin Riedmiller.
\newblock Playing {{Atari}} with {{Deep Reinforcement Learning}}, December
  2013.

\bibitem{schulmanProximalPolicyOptimization2017}
John Schulman, Filip Wolski, Prafulla Dhariwal, Alec Radford, and Oleg Klimov.
\newblock Proximal {{Policy Optimization Algorithms}}, August 2017.

\bibitem{qwen2}
An~Yang, Baosong Yang, Binyuan Hui, Bo~Zheng, Bowen Yu, Chang Zhou, Chengpeng
  Li, Chengyuan Li, Dayiheng Liu, Fei Huang, Guanting Dong, Haoran Wei, Huan
  Lin, Jialong Tang, Jialin Wang, Jian Yang, Jianhong Tu, Jianwei Zhang,
  Jianxin Ma, Jin Xu, Jingren Zhou, Jinze Bai, Jinzheng He, Junyang Lin, Kai
  Dang, Keming Lu, Keqin Chen, Kexin Yang, Mei Li, Mingfeng Xue, Na~Ni, Pei
  Zhang, Peng Wang, Ru~Peng, Rui Men, Ruize Gao, Runji Lin, Shijie Wang, Shuai
  Bai, Sinan Tan, Tianhang Zhu, Tianhao Li, Tianyu Liu, Wenbin Ge, Xiaodong
  Deng, Xiaohuan Zhou, Xingzhang Ren, Xinyu Zhang, Xipin Wei, Xuancheng Ren,
  Yang Fan, Yang Yao, Yichang Zhang, Yu~Wan, Yunfei Chu, Yuqiong Liu, Zeyu Cui,
  Zhenru Zhang, and Zhihao Fan.
\newblock Qwen2 technical report.
\newblock {\em arXiv preprint arXiv:2407.10671}, 2024.

\bibitem{Qwen2VL}
Peng Wang, Shuai Bai, Sinan Tan, Shijie Wang, Zhihao Fan, Jinze Bai, Keqin
  Chen, Xuejing Liu, Jialin Wang, Wenbin Ge, Yang Fan, Kai Dang, Mengfei Du,
  Xuancheng Ren, Rui Men, Dayiheng Liu, Chang Zhou, Jingren Zhou, and Junyang
  Lin.
\newblock Qwen2-vl: Enhancing vision-language model's perception of the world
  at any resolution.
\newblock {\em arXiv preprint arXiv:2409.12191}, 2024.

\bibitem{qwen2.5-VL}
Qwen Team.
\newblock Qwen2.5-vl, January 2025.

\end{thebibliography}

\end{document}